\documentclass[runningheads]{llncs}

\usepackage[T1]{fontenc}

\usepackage{graphicx}

\usepackage{color}

\usepackage{comment}

\usepackage{amsmath} 

\usepackage{booktabs}

\usepackage{tcolorbox}

\usepackage{nicefrac}

\usepackage{tikz}
\usetikzlibrary{positioning, shapes.geometric, arrows.meta}

\usepackage{pgfplots}
\pgfplotsset{compat=1.17}
\usepgfplotslibrary{groupplots}
\usepackage{sansmath}
\definecolor{stub}{HTML}{003F5C}
\definecolor{mid1}{HTML}{58508D}
\definecolor{mid2}{HTML}{BC5090}
\definecolor{rich}{HTML}{FF6361}

\begin{document}

\title{SQuIREL: A Context-Relative Framework for Measuring Entity Richness in Wikidata}

\titlerunning{SQuIREL: A Context-Relative Framework}

\author{Michelle Nwachukwu\and Albert Mero\~no-Pe\~nuela \and {Martim Brand\~ao}}

\authorrunning{M. Nwachukwu et al.}

\institute{King's College London, London, United Kingdom\\
\email{\{michelle.nwachukwu,albert.merono,martim.brandao\}@kcl.ac.uk}}

\maketitle

\begin{abstract}

Wikidata is a large, heterogeneous knowledge graph and applying data quality metrics requires context-specific approaches. The ``richness'' quality metric is currently under-researched and typically restricted to the ontology level, which fails to capture entity-specific structural depth. To address this gap, we introduce SQuIREL, a method that measures richness directly at the entity level using six dimensions: Semantic \textbf{S}pecificity, \textbf{Qu}alifier Density, \textbf{I}nternal Connectivity, \textbf{R}eference Provenance, \textbf{E}xternal Network Connectivity, and \textbf{L}inguistic Accessibility. These dimensions are kept disaggregated to preserve actionability for targeted editorial intervention. We apply this metric to two case studies of oil paintings (Brazilian vs. French) and writers (grouped by citizenship and gender) to demonstrate how it surfaces representational inequalities that may indicate systemic bias. Our metric successfully distinguishes between highly tailored and basic entity profiles, revealing nuances that raw counts and existing aggregate scores obscure. Specifically, it identifies structural imbalances, such as 48\% of Brazilian paintings clustering as low-richness stubs against 0\% in the French cohort, as well as intersectional patterns, including contrasting gender gaps between UK and Nigerian writers.

\keywords{Wikidata \and Bias \and Richness \and Data Quality.}
\end{abstract}

\section{Introduction}

Large-scale Knowledge Graphs (KGs) such as Wikidata serve as foundational infrastructure for modelling, storing, and reasoning over complex knowledge domains. Considerable attention has been paid to measuring KG quality, with dimensions such as completeness, accuracy, and consistency being prolific in the literature \cite{piscopo2019we,xue2022knowledge}. But Wikidata's immense scale and inherent heterogeneity make global, one-size-fits-all quality measurements highly ineffective. The underlying heterogeneity problem means that applying a uniform completeness metric across diverse entity classes ignores the varying structural and semantic requirements of different KG subgroups. For example, a planet entity and a film entity will have different modelling requirements; a single metric of data quality treating both alike is uninformative. While extensive literature exists to formalise dimensions of KG quality, work on adjacent concepts like ``richness'' remains limited, though the term is frequently used to highlight Wikidata's depth and interconnectivity \cite{colla2021wikidata,saez2018automatically,mbouadeu2023evaluation}. Where formalisations exist, they operate at the schema level, aggregating scores across an ontology or graph as a whole.

As Wikidata is increasingly utilised as a tool for AI systems, such as question answering, semantic search, and large language models, any biases in the graph could be inherited and amplified. If specific demographic cohorts, such as female scientists from the Global South, suffer from systemic representation bias or lack ``rich'' data due to lower editorial attention, downstream applications will structurally flatten these entities. Therefore, identifying and surfacing disparities goes beyond a graph maintenance task and becomes a critical requirement for ensuring equitable performance in the semantic web and AI ecosystem.

To operationalise a metric that allows such analyses of equity, this paper adapts the concept of richness to the entity level within Wikidata. We formally define a ``rich'' entity as one that features highly specific and well-referenced claims, utilises qualifiers extensively, acts as a central hub within Wikidata, connects broadly to external linked data, and is accessible across multiple languages. We propose SQuIREL, a context-specific richness metric framework that inherently accounts for the diverse requirements of different entity subgroups. We frame richness as complementary to completeness, acting as a separate quality pillar that captures depth and tailoring: where completeness asks whether expected properties are present, richness asks whether those properties carry the specificity, qualifiers, references, and multilingual accessibility that distinguish a tailored entity profile from a templated one.

We decompose richness into six specific dimensions following Wikidata's multi-layered structure. The framework reifies claims as statement nodes carrying qualifiers and references; complements the triple layer with multilingual labels, descriptions, and aliases; functions as a hub within the broader linked data graph; and acts as an index of external identifiers across institutional knowledge bases. Each of these is a distinct artifact that can be populated independently. Rather than collapse these into a single score, we keep the dimensions disaggregated to preserve the distinct editorial signal each carries, allowing interventions to target the specific features a cohort underuses.

In summary, we contribute the following: \textbf{(1)} a multi-dimensional, entity-level richness framework for Wikidata, grounded in the distinctive features of its data model and disaggregated by design to preserve the actionable signal each dimension carries; \textbf{(2)} an adaptation of Information Content theory from concepts to property-value pairs within a bounded cohort, supporting context-relative specificity scoring in heterogeneous knowledge graphs; \textbf{(3)} two case studies demonstrating the framework's utility as a tool for detecting representational disparity, including intersectional patterns that aggregate metrics obscure; and \textbf{(4)} an empirical validation against raw counts, sitelinks, Wikidata's Weighted Quality Score, and Recoin, showing that the framework captures signals these benchmarks miss.

\section{Background and Related Work}

\textbf{Wikidata Quality}
Foundational work by Lehmann et al. \cite{lehmann2016quality} on linked data quality presents 18 dimensions and 69 metrics, providing a common vocabulary for the field. Building on this, the first systematic literature review of data quality research in Wikidata applied a standard framework to group quality into categories, identifying completeness as by far the most researched, with calls for future work to produce reusable metrics \cite{piscopo2019we}. Subsequent work has shifted from cataloguing dimensions to detecting concrete failures. Indicators of low quality in Wikidata have been defined, such as outdated constraints, duplicate entities, and modelling errors \cite{shenoy2022study}. These indicators capture community consensus instead of abstract notions, moving beyond description-based surveys to provide actionable detection of specific problems. More recent work extends this perspective downstream, formalising what happens when constraint violations are repaired and showing how community bulk repairs have potential negative side-effects to the graph \cite{ferranti2025formalizing}. There is also research focused on references rather than statements. One framework extended typical reference dimensions, such as relevancy and authoritativeness, into 40 metrics across 22 dimensions, providing the first large-scale empirical characterisation of Wikidata reference quality across topical and random subsets \cite{hosseini2024rqss}. Source quality has also been examined across these same dimensions, alongside ease of access, in a multilingual evaluation \cite{amaral2021assessing}. These works characterise quality at the graph or statement level, but do not address how depth of representation varies across entity cohorts.

\textbf{Wikidata Completeness}
Completeness has been studied across various works in the literature \cite{prasojo2016managing,ahmeti2017assessing,darari2016enabling}, generally defined as the extent to which all required information is present in a dataset \cite{lehmann2016quality}. Consisting of sub-dimensions such as schema, property, population, and interlinking, approaches to measure completeness vary from testing against gold standards \cite{farber2017linked} to comparisons with human statements from contributors.
Recoin is a tool that offers a different angle by measuring completeness relative to peer entities instead of against an external reference \cite{balaraman2018recoin}. For each entity, Recoin identifies the top-five most common properties among similar entities that the target entity lacks, then computes completeness as the complement of those properties' average frequency. Recoin operates at the single-entity level and addresses property completeness specifically where its output is a list of likely-missing properties for one item at a time. Recoin can also compute the average frequency of the missing top five properties, mapping the average percentage of missing properties directly into 5 levels. Our framework is complementary, operating at the cohort level where one component of the framework scores representational depth of property-value pairs, and not just property presence.

\textbf{Richness}
Several structural quality frameworks in the literature evaluate KGs along dimensions that overlap with richness, such as fine-grained ontology usage, property utilisation, and class granularity, but stop short of defining richness as an explicit construct \cite{seo2022structural}. Foundational work on richness metrics focus on ontology schemas, proposing relationship, attribute, inheritance, and class richness metrics \cite{tartir2005ontoqa}. The knowledge base as a whole is made the unit of analysis, and the metrics are not context-specific. A subsequent framework explicitly defines properties, attribute, class, and annotation richness, but again applies these at the schema level and analyses the ontology as a whole, with no context or cohort specificity \cite{duque2011oquare}. Our framework differs in that it operates at the entity level and normalises within a bounded context, so that the same raw value can register as rich in one context and thin in another.

\textbf{Information Content}
Information Content (IC) was first introduced as a measure of the surprisal of an event \cite{shannon1948mathematical}. The notion was then adapted to concepts in an IS-A taxonomy for semantic similarity \cite{resnik1995using}, and further built on to define an information-theoretic similarity measure \cite{lin1998information}. Seco et al. \cite{seco2004intrinsic} subsequently proposed an intrinsic IC formulation that derives concept specificity from taxonomic structure alone, removing the dependence on corpus statistics. These ideas were extended to the Semantic Web \cite{meymandpour2016semantic}, and generalised IC to the Resource Description Framework (RDF).
More recent frameworks operationalise IC-based similarity over KGs, including general-purpose toolkits such as Sematch \cite{zhu2017sematch} and supervised, aspect-oriented approaches for biomedical KGs \cite{sousa2022supervised}. In this paper we extend IC from concepts to property-value pairs within a cohort, with weighting by property prevalence.

\textbf{Wikidata Bias}
Despite its mission as a global, multilingual resource, Wikidata is characterised by persistent structural inequalities. Demographic and geographic skews among editors and funders \cite{wikidatasurvey2021,turki2023ten} result in content disparities that over-represent the Global North and male entities across diverse domains \cite{shaik2021analyzing,wagner2015s,abian2022analysis}. While these biases are well-documented, the literature lacks an entity-level, context-relative instrument to surface specific underdeveloped dimensions within these cohorts. We therefore treat uneven metric scores as representational disparity rather than direct editor bias, acknowledging that these results also reflect external source availability and domain-specific modelling conventions.

\section{Preliminaries}

\subsection{Wikidata Structure}

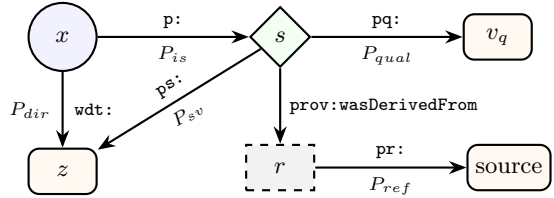
\begin{figure}[htpb]
    \centering
    \begin{tikzpicture}[
        >=Stealth,
        item/.style={circle, draw, minimum size=0.9cm, thick, fill=blue!5, inner sep=1pt},
        stmt/.style={diamond, aspect=1.3, draw, minimum size=0.8cm, thick, fill=green!5, inner sep=1pt},
        val/.style={rectangle, draw, minimum height=0.6cm, minimum width=0.9cm, thick, rounded corners, fill=orange!5, inner sep=3pt},
        ref/.style={rectangle, draw, minimum height=0.6cm, minimum width=0.9cm, thick, dashed, fill=gray!10, inner sep=3pt},
        edge label/.style={midway, above, font=\scriptsize\ttfamily, sloped},
        edge label math/.style={midway, below, font=\scriptsize, sloped}
        ]

        \node[item] (item) {$x$};
        \node[stmt] (stmt) [right=2cm of item] {$s$};

        \node[val] (valdir) [below=1cm of item] {$z$};

        \node[val] (qualval) [right=2cm of stmt] {$v_q$};

        \node[ref] (refnode) [below=1cm of stmt] {$r$};

        \node[val] (srcenode) [right=2cm of refnode] {$\text{source}$};

        \draw[->, thick] (item) -- (valdir)
            node[midway, right=1pt, font=\scriptsize\ttfamily] {wdt:}
            node[midway, left=1pt, font=\scriptsize] {$P_{dir}$};

        \draw[->, thick] (item) -- (stmt)
            node[edge label] {p:}
            node[edge label math] {$P_{is}$};

        \draw[->, thick] (stmt) -- (valdir)
            node[edge label] {ps:}
            node[edge label math] {$P_{sv}$};

        \draw[->, thick] (stmt) -- (qualval)
            node[edge label] {pq:}
            node[edge label math] {$P_{qual}$};

        \draw[->, thick] (stmt) -- (refnode)
            node[midway, right=0.1pt, font=\scriptsize\ttfamily] {prov:wasDerivedFrom};

        \draw[->, thick] (refnode) -- (srcenode)
            node[edge label] {pr:}
            node[edge label math] {$P_{ref}$};

    \end{tikzpicture}
    \caption{How Wikidata models claims. The same fact can be expressed as a direct triple $(x, p, z)$ with $p \in P_{dir}$, or through a statement node $s \in S_G$ which additionally supports qualifiers and references. Edges are labelled with their RDF namespace and corresponding property subset.}
    \label{fig:wikidata_schema}
\end{figure}

Wikidata extends RDF and uses a statement-based data model to represent complex, verifiable facts \cite{malyshev2018getting}. Let $G$ denote the full Wikidata RDF graph. We define the set of core Wikidata items as $I_G=\{x \mid (x, \text{rdf:type, wikibase:Item}) \in G \}$, restricting to instances of $\texttt{wikibase:Item}$ to exclude internal nodes such as qualifiers and references. The set of all values in the graph is $V_G= \{z \mid \exists x, y:(x,y,z) \in G \}$.
Properties in Wikidata serve various structural functions. Direct Properties $(P_{dir})$ in the $\texttt{wdt:}$ namespace link items directly to values, with a subset $P_{ext} \subset P_{dir}$ representing external identifiers.
Complex claims (for example, facts constrained by temporal validity or supported by source references) are reified as statement nodes: $S_G = \{ s \mid (s,\text{ rdf:type, wikibase:Statement}) \in G \}$. An Item-to-Statement property $(P_{is})$ in the \texttt{p:} namespace connects an item $x \in I_G$ to a statement node $s$, and a Statement-to-Value property $P_{sv}$ in the \texttt{ps:} namespace connects s to its target value $z \in V_G$.
Statement nodes can further be enriched with qualifiers and references. Qualifier Properties $(P_{qual})$ in the \texttt{pq:} namespace link a statement to a qualifier value. Reference Nodes are defined as $R_G = \{r \mid (r, \text{ rdf:type, wikibase:Reference}) \in G \}$, attached to statements via the \texttt{prov:wasDerivedFrom} predicate. Reference Properties $(P_{ref})$ in the \texttt{pr:} namespace link a reference node to a source value.

\subsection{Context Definition}

Our richness metric is context-relative; scores are computed against a cohort of comparable items opposed to the graph as a whole. 
A \textit{Context Definition} $\Delta$ is a finite set of property-value pairs using direct properties,
$\Delta = \{(p_1, v_1), (p_2, v_2), ... , \\(p_i, v_i)\}$, where $p_i \in P_{dir}$ and $v_i \in V_G$.
The corresponding \textit{Context} $\mathcal{C}_{\Delta}$ is the subset of $I_G$ satisfying all pairs in $\Delta$:

\begin{equation*}
    \mathcal{C}_{\Delta} = \{ x \in I_G \mid \forall (p, v) \in \Delta : (x, p, v) \in G \}
\end{equation*}

For example, $\Delta = \{\texttt{(P31, Q5), (P106, Q36180)}\}$ defines the context of all humans whose occupation is writer.
For comparative analysis, we partition a context along additional properties. 
For instance, partitioning the writers context above by \texttt{P27} (country\_of\_citizenship) with \texttt{{Q1033, Q145}} yields Nigerian and UK writer sub-contexts; further partitioning each by \texttt{P21} (sex\_or\_gender) with \texttt{{Q6581072, Q6581097}} yields the four cohorts compared in Section 5.2.

\section{Formal Definitions of Metrics}

Building on our definition of entity richness, we formalise SQuIREL across six dimensions: Semantic Specificity, Qualifier Density, Reference Provenance, Internal Connectivity, External Network Connectivity, and Linguistic Accessibility. Each dimension captures a distinct aspect of entity development that matters in Wikidata. The dimensions are kept disaggregated rather than collapsed into a composite score, so editors can target the specific dimension where a cohort is underdeveloped.

\subsection{Dimension 1: Semantic Specificity}

This dimension is designed under the assumption that under-addressed entities often have only basic properties filled out (e.g., instance of, date of birth, sex or gender) or lack specific information. We reward entities with rare values that hold specific information to differentiate between entities that have templated representation from individual, tailored representation.  
Let $P_{sem}=P_{dir} \setminus P_{ext}$ denote all properties that are not external identifiers. We write $PV(x)$ for the set of $(p,v)$ pairs with $p \in P_{sem}$ asserted on $x$, and $PV(\mathcal{C}_{\Delta})$ for the union of such pairs across the context, counting each statement separately when an item has multiple values for $p$.
Let $N(p) = | \{(x,p,v) \in G \mid x \in \mathcal{C}_{\Delta}\}|$ be the absolute frequency of $p$ in the context, and $N(p,v) = | \{x \in \mathcal{C}_{\Delta} \mid (x,p,v) \in G\}|$ the absolute frequency of a property-value pair $(p,v)$. We define a prevalence weighting factor $\rho(p) = {N(p)}/{|\mathcal{C}_{\Delta}|}$. This factor does not depend on the individual item; it is intentionally calculated at the context level to measure how commonly a specific property is used across the entire context, rewarding items possessing rare values for common properties.
Following Shannon \cite{shannon1948mathematical}, we score each $(p,v)$ pair based on the Information Content (IC), representing the rarity of $v$ given $p$:

\begin{equation*}
    IC_{w}(p,v) = \rho(p) \cdot -\log_2\left(\frac{N(p,v)}{N(p)}\right)
\end{equation*}

To score an item $x$, we first calculate the mean of its weighted IC scores. Then, in a distinct, subsequent step, this mean is explicitly multiplied by a log-scaled breadth factor, which is derived from the item's total number of property-value pairs, rewarding entities that are both individually specific and broadly described:

\begin{equation*}
    \text{Spec}(x) = \frac{1}{|PV(x)|} \sum_{(p,v)\in PV(x)} IC_{w}(p,v) \cdot \log_{10}(1 + |PV(x)|)
\end{equation*}

We define the contextual maximum $\text{Spec}_{max}(\mathcal{C}_{\Delta}) = \max_{(p,v) \in PV(\mathcal{C}_{\Delta})}IC_{w}(p,v)$, and rescale $\text{Spec}(x)$ for a score interpretable as a ratio of the most informative pair: $\hat{\text{Spec}}(x) = \nicefrac{\text{Spec}(x)}{\text{Spec}_{max}(\mathcal{C}_{\Delta})}$.

\subsection{Dimension 2: Qualifier Density}

This dimension rewards the use of qualifiers, which adds nuance to an entity's representation (e.g. start time, end time, academic degree), and is a major indicator of editor attention and entity richness.

Let $P_{allowed}$ be the set of properties that structurally permit qualifiers, defined operationally as those passing the allowed qualifiers constraint (\texttt{Q21510851}) via property constraint \texttt{P2302}. 
The set of qualifiable statement nodes for an item $x$ is
$S_{q}(x) = \{s \in S_G \mid \exists p \in P_{allowed} : (x,p,s) \in G\}$. 
Similarly we define the set of qualifiable statement nodes across all items within the context, $\mathcal{C}_{\Delta}$: $S_{q}(\mathcal{C}_{\Delta}) := \{s \in S_G \mid \exists x \in \mathcal{C}_{\Delta},  p \in P_{allowed} : (x,p,s) \in G\}$. 
For each $s\in S_q(x)$, let $N_q(s) = \vert \{q \in P_{qual} \mid \exists v \in V_G: (s,q,v) \in G\} \vert$ count the distinct qualifiers attached to $s$.
The absolute Qualifier Density for an item $x$ is defined as the mean number of distinct qualifiers used per qualifiable statement:

\begin{equation*}
    \text{Dens}(x) = \left\{ \begin{array}{c r} 
    0 & \mbox{if }\vert S_q(x) \vert = 0 \\ 
    \frac{1}{\vert S_q(x) \vert} \sum_{s\in S_q(x)} N_q(s) & \mbox{otherwise} 
    \end{array} \right.
\end{equation*}

To robustly cap the extreme outliers inherent to Wikidata's scale-free distribution, we normalise by the 99th percentile across the context, $\text{Dens}_{99}(\mathcal{C}_{\Delta}) = P_{99}(\{N_q(s) \mid s \in S_q(\mathcal{C}_{\Delta})\})$.
Finally the normalised Qualifier Density score for an item $x$ is calculated as the ratio of its density to the 99th percentile, capped at 1 to account for the top 1\% of outliers: 
$\hat{\text{Dens}}(x) = \min \left(1,\frac{\text{Dens}(x)}{\text{Dens}_{99}(\mathcal{C}_{\Delta})}\right)$.

\subsection{Dimension 3: Reference Provenance}

This dimension rewards entities with documented provenance that align with Wikidata's referencing policy\footnote{\url{https://www.wikidata.org/wiki/Help:Sources}}.
We measure both the density and diversity of references, explicitly filtering out cross-wiki imports that violate the policy\ (e.g., \texttt{P143}: imported from Wikimedia project).

Let $S(x) = \{s \in S_G \mid \exists p \in P_{is} : (x,p,s) \in G\}$ be the set of statement nodes for $x$.
For any statement $s \in S(x)$, let $R_{G}(s)$ be the set of all reference nodes attached to $s$. We define $p_{import}$ as the property denoting a cross-wiki import (specifically \texttt{P143}). We isolate the set of valid references, $R_{valid}(s)$, by excluding any reference node that uses the import property, $p_{import} = \texttt{P143}$: $R_{valid}(s) = \{r \in R_G(s) \mid \neg \exists v \in V_G : (r,p_{import}, v) \in G\}$.

The total reference count and unique reference count for $x$ are, $N_{ref}(x) = \sum_{s\in S(x)} \vert R_{valid}(s) \vert$ and $U_{ref}(x) = \left|\bigcup_{s \in S(x)} R_{valid}(s)\right|$ respectively. 
The total reference count measures density by counting every single citation attached to every statement on an item. The unique reference count measures diversity by looking at the item as a whole and counting the number of distinct, individual sources used.
The Reference Provenance score equally weights density and diversity, normalised by the number of statements. We weight density and diversity equally, treating both as primary indicators of provenance quality. Since a high-quality baseline is one unique reference per statement, we cap the maximum at 1:

\begin{equation*}
    \text{Prov(x)} = \left\{ \begin{array}{c r} 
    0 & \mbox{if }\vert S(x) \vert = 0 \\ 
    \min \left(1,\frac{0.5 \cdot N_{ref}(x) + 0.5 \cdot U_{ref}(x)}{\vert S(x) \vert} \right) & \mbox{otherwise} 
    \end{array} \right.
\end{equation*}

\subsection{Dimension 4: Internal Connectivity}

This dimension evaluates in-degree centrality and measures how deeply integrated an entity is within the broader knowledge graph. Because of the scale-free nature of Wikidata, we apply logarithmic scaling to prevent extreme hub nodes from skewing the metric.

The absolute in-degree of $x$ is the number of distinct item-property pairs $(e,p)$ where another entity links to $x$, via a direct property, $deg_{in}(x) = \vert \{(e,p) \mid e \in I_G \wedge p \in P_{dir} \wedge (e,p,x)\in G \} \vert$.
We apply a base-10 logarithmic transformation, $L_{in}(x) = \log_{10}(1+deg_{in}(x))$, and normalise by the 99th percentile across the context, $L_{99}(\mathcal{C}_{\Delta}) = P_{99} (\{L_{in}(y) \mid y \in \mathcal{C}_{\Delta}\})$, capped at 1: $\hat{\text{Conn}(x)} = \min \left( 1, \frac{L_{in}(x)}{L_{99}(\mathcal{C}_{\Delta})} \right)$.

\subsection{Dimension 5: External Network Connectivity}

This dimension utilises external identifiers as evidence of institutional recognition, highlighting whether the broader semantic web and traditional academic or cultural institutions have recognised the entity, a factor heavily influenced by systemic biases. 

The absolute external connectivity is the number of distinct external identifiers associated with $x$, $N_{ext}(x) = \vert \{p \in P_{ext} \mid \exists v \in V_G : (x,p,v) \in G\} \vert$.
We normalise by the 99th percentile across the context, $\text{Ext}_{99}(\mathcal{C}_{\Delta}) = \mbox{P}_{99}(\{N_{ext}(y) \mid y \in \mathcal{C}_{\Delta}\})$, capped at 1: $\hat{\text{Ext}}(x) = \min \left(1,\frac{N_{ext}(x)}{\text{Ext}_{99}(\mathcal{C}_{\Delta})}\right)$.

A high average score across the context indicates that the items are broadly integrated into similar external databases and authority control systems.

\subsection{Dimension 6: Linguistic Accessibility}

This dimension evaluates Wikidata's multilingualism, specifically whether it lacks the comprehensive translations needed to make the graph accessible to diverse linguistic communities. We measure accessibility via the presence and quality of textual properties.
Let $\Sigma^*$ denote the set of all valid character strings, and $\Lambda$ denote the set of evaluated languages. We define $\Lambda$ context-relatively as the set of languages that appear in at least one label across $\mathcal{C}_{\Delta}$, so that the metric evaluates accessibility against the languages actually represented in the cohort and not a fixed global set. We consider three attributes: 

\begin{itemize}
    \item \textbf{Labels:} a unique name for an item in a language, defined as a partial function, $\lambda: I_G \times \Lambda \to \Sigma^*$.

    \item \textbf{Descriptions:} a short summary of an item in a language, $\delta: I_G \times \Lambda \to \Sigma^*$.

    \item \textbf{Aliases:} alternative names for an item in a language, mapped to the power set of strings, $\alpha: I_G \times \Lambda \to \mathcal{P}(\Sigma^*)$.
\end{itemize}

\noindent We score alias accessibility for $x$ in language $l$ with a logistic sigmoid. We apply a sigmoid function to measure aliases so as to heavily reward the first synonyms but prevent a long list of synonyms from artificially inflating the metric:

\begin{equation*}
    \text{Acc}_{\alpha}(x,l) = \left\{ \begin{array}{l l} 0 & \mbox{if } |\alpha(x,l)| = 0 \\\frac{1}{1 + e^{-(|\alpha(x,l)|-1)}} & \mbox{otherwise}
    \end{array}\right.
\end{equation*}

For description accessibility, we follow Wikidata's official guidelines \footnote{\url{https://www.wikidata.org/wiki/Help:Description}}, which encourage brevity. Let $M_{l}$ be the median character length of all descriptions in language $l$. We penalise descriptions that exceed either 100 characters or twice the language median:

\begin{equation*}
    \text{Acc}_{\delta}(x,l) = \left\{ \begin{array}{l l} \max \left( 0,1 - \frac{\max(0, |\delta(x,l)|-\max(100, 2M_l))}{\max(100, 2 M_l)} \right) & \mbox{if } \delta(x,l) \mbox{ is defined} \\ 0   & \mbox{otherwise}
    \end{array}\right.
\end{equation*}

The overall linguistic accessibility for $x$ in a language $l$ is the weighted sum of its components, with the score set to 0 if the primary label is missing:

\begin{equation*}
    \text{Acc}_{\Lambda}(x,l) = \left\{ \begin{array}{l l}
     0.6 + 0.3 \cdot \text{Acc}_{\delta}(x,l) + 0.1 \cdot \text{Acc}_{\alpha}(x,l)  & \mbox{if } \lambda(x,l) \mbox{ is defined} \\
      0   & \mbox{otherwise}
    \end{array}\right.
\end{equation*}

The component weights reflect a hierarchy of accessibility and were not empirically calibrated. The label carries the largest weight (0.6) as the minimum requirement for an entity to be reachable in a language; descriptions (0.3) provide contextual comprehension but are not required for retrieval; aliases (0.1) improve recall and accommodate naming variants without affecting baseline access. 
The absolute Linguistic Accessibility score is summed across all evaluated languages: $\text{Lang}(x) = \sum_{l \in \Lambda} Acc_{\Lambda}(x,l)$.

Given a theoretical maximum of $\text{Lang}_{max}(\Lambda) = |\Lambda|$ (the per-language ceiling $0.6 + 0.3 + 0.1 = 1$), the score normalises to: $\hat{\text{Lang}}(x) = \nicefrac{\text{Lang}(x)}{\text{Lang}_{max}(\Lambda)}$.

\section{Case Studies}

We consider two case studies centred around comparing richness across two cohorts: 1) Brazilian vs French oil paintings, and 2) UK versus Nigerian writers. These case studies were selected specifically to include one comparison with comparable cohort sizes and one with vastly different sizes, demonstrating the framework’s utility regardless of baseline volume. We used QLever \cite{bast2017qlever} to extract entity data via SPARQL queries, computed all six richness dimensions for each entity in each cohort, and analysed the resulting distributions. 

\textbf{Detecting Disparities} We treat inequality in metric scores as evidence of representational disparity, which may reflect editor attention or other factors as discussed in Section 2.
Instead of absolute thresholds, we rely on between-cohort comparison, distinguishing underlying disparities 
from cases where a metric is less applicable to a domain. Statistical comparisons use Mann-Whitney U for two-cohort tests and Kruskal-Wallis H for multi-cohort tests. 

\textbf{Origins of Representational Disparity} While unequal metric scores expose disparity, the metric alone cannot automatically distinguish the root cause. As detailed in the Documentation Framework in the appendix, we categorise observed disparities into three distinct root causes:

\begin{itemize}
    \item Editorial Gap: Disparities caused by a skew in editor attention or behaviour.
    \item Source Material Gap: Disparities caused by a lack of verifiable references and source material, reflecting external realities rather than internal biases.
    \item Data Model Gap: Disparities caused by a mismatch between Wikidata’s established schema and the epistemic reality of the target context.
\end{itemize}

Identifying which of these mechanisms drives lower richness requires complementary investigation, though SQuIREL provides the necessary signal to begin that process.

\textbf{Cross-dimension Comparison} Using our richness metric, each dimension is normalised independently against its own contextual benchmark, so scores are comparable across cohorts within a dimension but not directly across dimensions: a Specificity of 0.3 and a Provenance of 0.3 reflect different underlying quantities. Figures \ref{fig:painting_distri} and \ref{fig:writer_distri} display the within-cohort shape of each dimension and support cross-cohort comparison along each dimension.

\subsection{Case Study 1: French and Brazilian Oil Paintings}

Our base context is $\Delta = \{\texttt{(P31, Q3305213), (P186, Q296955)}\}$ (paintings made from oil paint), grouped by \texttt{P495} (country of origin) into \texttt{Q155} Brazil and \texttt{Q142} France, consisting of 852 and 901 paintings respectively.

\begin{figure*}[t]
\centering
\includegraphics[width=0.8\textwidth]{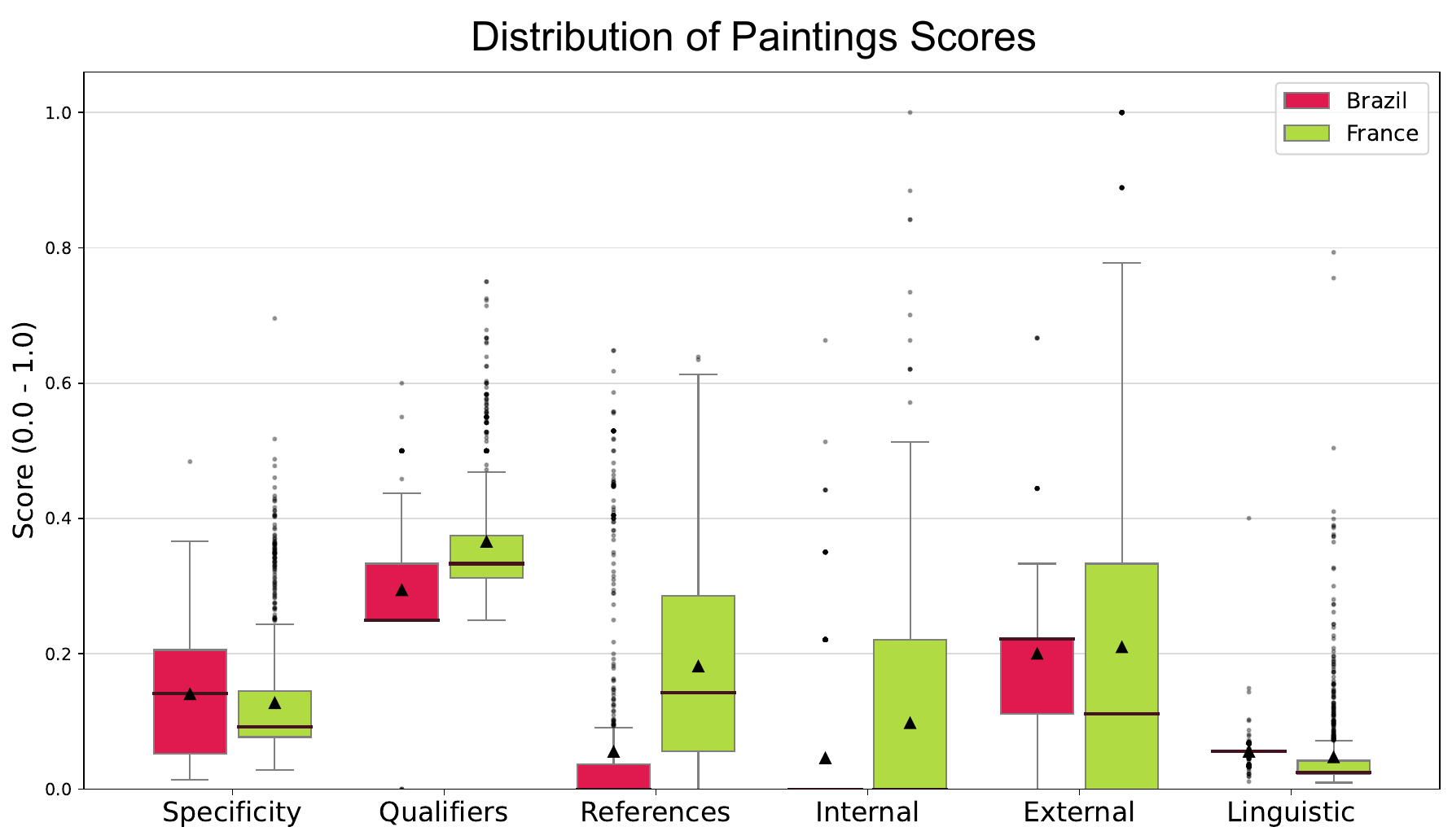} 
\caption{Normalised score distributions for Brazilian and French oil painting entities.}
\label{fig:painting_distri}
\end{figure*}

\textbf{Overall Scores}
Figure \ref{fig:painting_distri} shows the distribution of normalised scores across the six dimensions. The most pronounced difference occurs in References and Internal Connectivity, where Brazilian paintings cluster near zero while French paintings exhibit broad variance, consistent with editorial attention and graph integration concentrated on the French cohort. France also shows higher medians and broader variance on Qualifier Density. Brazilian paintings show higher medians on Specificity, External Connectivity, and Linguistic Accessibility, though France's External distribution has a wider tail and slightly higher mean, indicating that a smaller subset of French paintings achieves much deeper external integration. Linguistic Accessibility shows a similarly large effect ($r = 0.60$), but with Brazilian paintings scoring higher than French. All six dimensions differ significantly between cohorts (Mann-Whitney U, $p < 0.001$).

\textbf{Grouped Comparisons} 
We clustered entities by their six-dimensional richness profile to distinguish rich from stub items, using a Gaussian Mixture Model rather than K-Means. GMM estimates the covariance structure of the data, accommodating elliptical and variably-dense clusters that K-Means cannot. The optimal number of components, $k = 4$, was selected via Bayesian Information Criterion.

\begin{figure}[htbp]
\centering
{\sffamily\sansmath
\begin{tikzpicture}
\begin{axis}[
    ybar stacked,
    width=\linewidth,
    height=0.75\linewidth,
    bar width=0.8,
    xmin=-0.5, xmax=1.5,
    ymin=0, ymax=100,
    xtick={0,1},
    xticklabels={Brazil,France},
    ytick={0,20,40,60,80,100},
    tick label style={font=\footnotesize},
    ylabel={Percentages (\%)},
    title={Stacked Bar Chart of Paintings Across Two Categories},
    legend style={
        at={(0.99,0.01)}, anchor=south east,
        font=\scriptsize, draw=black!20,
        fill=white, fill opacity=0.85, text opacity=1,
        cells={anchor=west},
    },
    legend image code/.code={\fill[#1] (0cm,-0.1cm) rectangle (0.4cm,0.1cm);},
]
\addplot[fill=stub, draw=none] coordinates {(0,48) (1,0)};
\addplot[fill=mid1, draw=none] coordinates {(0,38) (1,62)};
\addplot[fill=mid2, draw=none] coordinates {(0,1)  (1,18)};
\addplot[fill=rich, draw=none] coordinates {(0,13) (1,20)};
\legend{Stub Entities, Mid-Range 1,
        Mid-Range 2, Rich Entities}
\coordinate (L0) at (axis cs:0,24);   \coordinate (L1) at (axis cs:0,67);
\coordinate (L2) at (axis cs:0,86.5); \coordinate (L3) at (axis cs:0,93.5);
\coordinate (L4) at (axis cs:1,0);    \coordinate (L5) at (axis cs:1,31);
\coordinate (L6) at (axis cs:1,71);   \coordinate (L7) at (axis cs:1,90);
\end{axis}
\begin{scope}[every node/.style={white, font=\footnotesize}]
    \node at (L0) {48\%}; \node at (L1) {38\%};
    \node at (L2) {1\%};  \node at (L3) {13\%};
    \node at (L4) {0\%};  \node at (L5) {62\%};
    \node at (L6) {18\%}; \node at (L7) {20\%};
\end{scope}
\end{tikzpicture}}
\caption{Proportion of painting entities by richness category in Brazil and France.}
\label{fig:painting_stacked}
\end{figure}
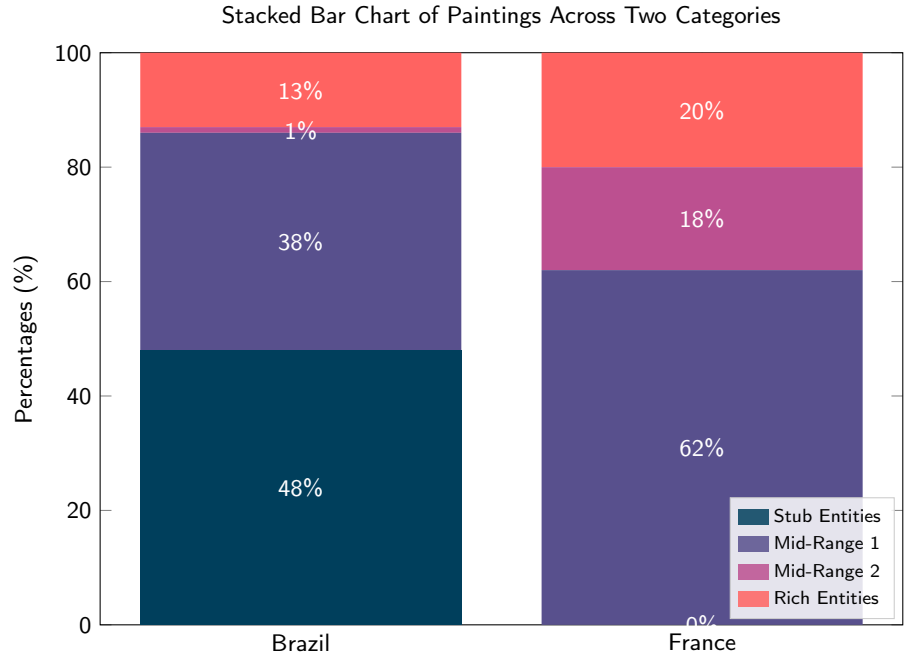

Figure \ref{fig:painting_stacked} shows the proportion of painting entities by richness category. The resulting cluster composition by cohort are shown in the appendix. \textbf{Low Richness Stub entities} are exclusively Brazilian (48\% of the Brazilian cohort, 0\% of French), characterised by near-complete absence of Internal Connectivity and References, paired with limited Qualifiers and External links. \textbf{Mid-range cluster 1} combines low Internal Connectivity with moderate External links (38\% of Brazilian, 62\% of French). \textbf{Mid-range cluster 2} has high Internal Connectivity but fewer External links and is rare among Brazilian paintings (1\%) but more common among French (18\%); both mid-range clusters show high Qualifier density. \textbf{Rich entities} show the highest scores across nearly all dimensions and are the only cluster with elevated Specificity and Linguistic scores, suggesting attention to detail in curation; they account for 13\% of Brazilian and 20\% of French paintings. Kruskal-Wallis H tests confirm significant differences across all six dimensions between the four clusters, and a Chi-Square test of independence confirms that cluster membership depends strongly on cohort $(p < 0.001)$.

\textbf{Gini Scores} 
We also computed Gini scores for each cohort, as a way of measuring inequality entity richness within each cohort. The full Gini scores for both case studies are provided in the appendix. 
Both cohorts show low Gini scores for Specificity and Qualifiers, indicating relatively even distributions, but diverge in other dimensions. A distinct inequality is in References, where Brazil's score (0.85) reflects a near-binary split between paintings with no references and a small group with several. France's higher External Gini (0.57) is consistent with a smaller subset of high-profile paintings accumulating disproportionate external links, a natural feature of a cohort containing internationally recognised works. Internal Connectivity Gini is high in both cohorts, which we attribute to reflecting Wikidata's scale-free linked structure and not a cohort-specific imbalance.

These results indicate that French painting entities are more deeply integrated, whereas nearly half of the Brazilian cohort are low-richness stubs, identifying key areas for future curation.

\subsection{Case Study 2: Writers Across Citizenship and Gender}

Our base context is $\Delta = \{\texttt{(P31, Q5), (P106, Q36180)}\}$ (humans whose occupation is writer), grouped by \texttt{P27} (country of citizenship) into \texttt{Q1033} Nigerian and \texttt{Q145} UK, then further grouped by \texttt{P21} (sex or gender) into \texttt{Q6581072} female and \texttt{Q6581097} male. Cohort sizes are 426 (Nigerian female), 1,396 (Nigerian male), 4,649 (UK female), and 7,996 (UK male).

\begin{figure*}[t]
\centering
\includegraphics[width=0.9\textwidth]{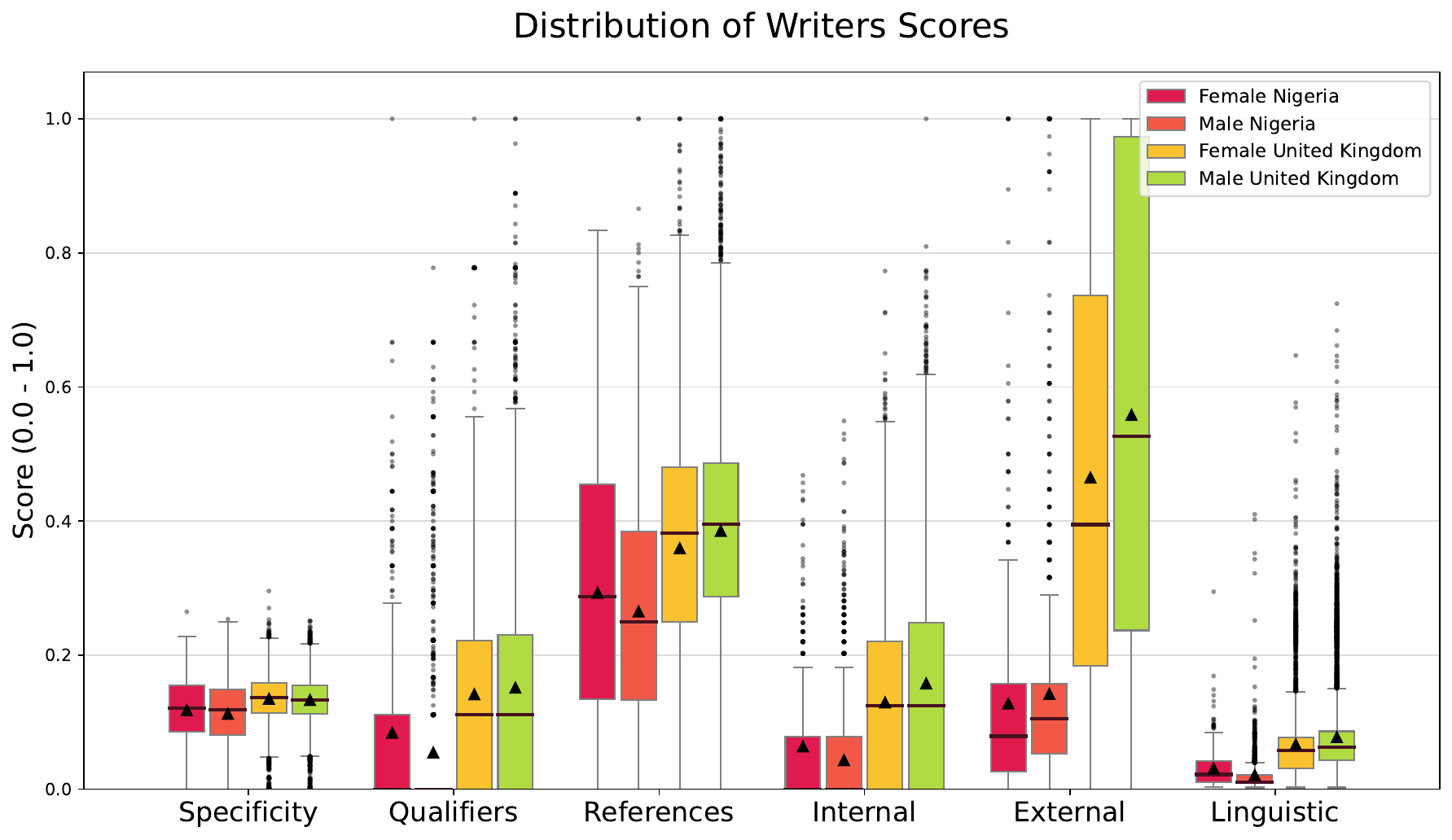} 
\caption{Normalised score distributions for writer entities by gender and citizenship. Overall differences across groups were evaluated using a Kruskal-Wallis omnibus test}
\label{fig:writer_distri}
\end{figure*}

\textbf{Overall Scores}
Citizenship is the primary driver of variance (Figure \ref{fig:writer_distri}), where UK writers score significantly higher than Nigerian writers across all six dimensions (all $p < 0.001$), with the largest gap in External Connectivity and UK males in particular exhibiting a vastly wider interquartile range. Gender effects are smaller and direction-dependent. In the Nigerian cohort, female writers score slightly higher than male in References, with no significant Specificity difference ($p = 0.396$). One possible explanation is that Nigerian female writers sit at an intersection that has received targeted editorial attention from debiasing initiatives meaning the smaller cohort that exists has been more deliberately curated. In the UK cohort, male writers score significantly higher in References, External, and Linguistic dimensions, while female writers show a slight but significant Specificity advantage; no significant difference was found in Qualifiers ($p = 0.129$). Specificity and Linguistic distributions are tighter and lower across all four cohorts. Full pairwise comparisons are provided in the appendix.

\textbf{Grouped Comparisons}
BIC selected $k = 3$ for this context. \textbf{Low Richness Stub entities} average zero on Qualifier score and the lowest scores across all six dimensions; the majority of Nigerian writers fall into this cluster (66\% female, 80\% male). \textbf{Mid entities} show large gains in External, References, and Qualifier scores, accounting for just over half of UK female writers (53\%) and 44\% of UK male; this category is much larger in the UK cohort than in Nigeria. \textbf{Rich entities} show the highest scores across all dimensions and a dramatically high External score; only 1\% of both Nigerian female and Nigerian male writers fall here, compared to 15\% of UK female and 24\% of UK male.

\begin{figure}[htbp]
\centering
{\sffamily\sansmath
\begin{tikzpicture}
\begin{axis}[
    ybar stacked,
    width=\linewidth,
    height=0.75\linewidth,
    bar width=0.8,
    xmin=-0.5, xmax=3.5,
    ymin=0, ymax=100,
    xtick={0,1,2,3},
    xticklabels={Female Nigeria, Male Nigeria, Female United Kingdom, Male United Kingdom},
    ytick={0,20,40,60,80,100},
    tick label style={font=\footnotesize},
    x tick label style={font=\scriptsize},
    ylabel={Percentages (\%)},
    title={Stacked Bar Chart of Writers Across Four Categories},
    legend style={
        at={(0.99,0.99)}, anchor=north east,
        font=\scriptsize, draw=black!20,
        fill=white, fill opacity=0.85, text opacity=1,
        cells={anchor=west},
    },
    legend image code/.code={\fill[#1] (0cm,-0.1cm) rectangle (0.4cm,0.1cm);},
]
\addplot[fill=stub, draw=none] coordinates {(0,66) (1,80) (2,32.5) (3,32)};
\addplot[fill=mid1, draw=none] coordinates {(0,33) (1,19) (2,52.5) (3,44)};
\addplot[fill=mid2, draw=none] coordinates {(0,1)  (1,1)  (2,15)   (3,24)};
\legend{Stub Entities, Mid-Range, Rich Entities}
\coordinate (A0) at (axis cs:0,33);   \coordinate (A1) at (axis cs:0,82.5); \coordinate (A2) at (axis cs:0,99.5);
\coordinate (B0) at (axis cs:1,40);   \coordinate (B1) at (axis cs:1,89.5); \coordinate (B2) at (axis cs:1,99.5);
\coordinate (C0) at (axis cs:2,16.25);\coordinate (C1) at (axis cs:2,58.75);\coordinate (C2) at (axis cs:2,92.5);
\coordinate (D0) at (axis cs:3,16);   \coordinate (D1) at (axis cs:3,54);   \coordinate (D2) at (axis cs:3,80.5);
\end{axis}
\begin{scope}[every node/.style={white, font=\footnotesize}]
    \node at (A0) {66\%}; \node at (A1) {33\%}; \node at (A2) {1\%};
    \node at (B0) {80\%}; \node at (B1) {19\%}; \node at (B2) {1\%};
    \node at (C0) {33\%}; \node at (C1) {53\%}; \node at (C2) {15\%};
    \node at (D0) {32\%}; \node at (D1) {44\%}; \node at (D2) {24\%};
\end{scope}
\end{tikzpicture}}
\caption{Proportion of writer entities by richness category across gender and citizenship.}
\label{fig:writer_stacked}
\end{figure}
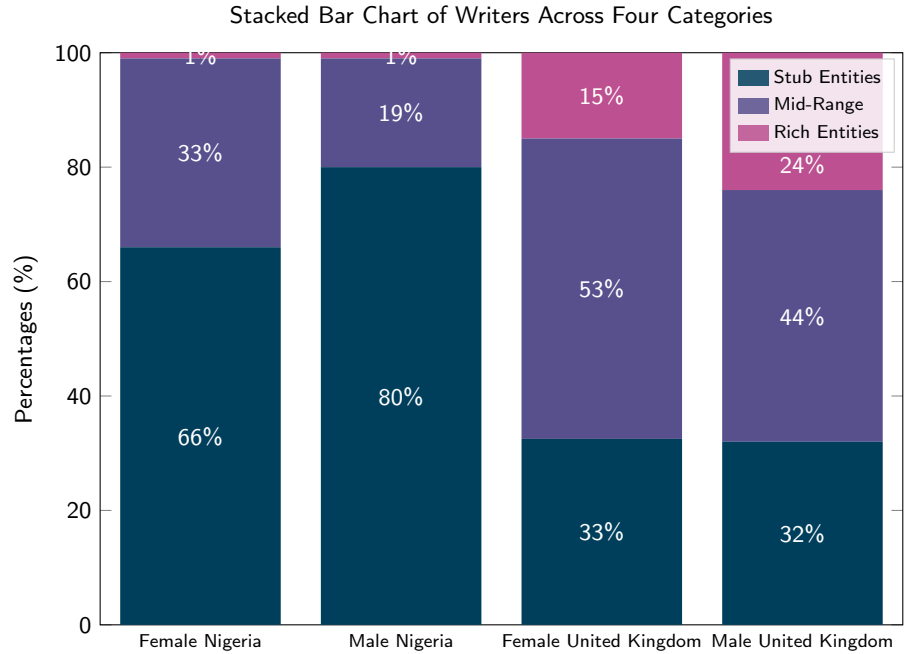

\textbf{Gini Scores}
Internal inequality is highest in the Qualifier and Internal dimensions across all four cohorts, with Nigerian male writers showing the most extreme imbalance (0.86 Qualifier, 0.79 Internal). Female Nigerian writers fall closely behind. The Linguistic dimension is the only one with comparable Gini scores across all four cohorts. UK Gini scores are consistently lower than Nigerian scores across Specificity, Qualifiers, References, and Internal, suggesting that the unevenness in the Nigerian cohort reflects representational disparity rather than a natural feature of the data.

These results reveal that citizenship outweighs gender in determining profile richness, while also exposing unexpected characteristics that give editors a deeper understanding of Wikidata.

\section{Validation Against Existing Quality Indicators}

To further validate the proposed framework, we examined whether the six dimensions capture properties beyond what existing quality indicators already provide. We compared each dimension against four classes of existing indicators: simple structural counts (triples, qualifiers, references, languages), Wikipedia integration (sitelinks), Wikidata's Weighted Quality Score (WQS), a widely-used aggregate quality tool on Wikidata\footnote{\url{https://item-quality-evaluator.toolforge.org/}}, and Recoin \cite{balaraman2018recoin}. Spearman's rank correlation was used to accommodate non-linear monotonic relationships and the non-normal distribution of the underlying scores. We computed correlations separately for each cohort to surface context-dependent behaviour. The full correlation matrix is provided in the appendix.

\subsection{Reference and Weighted Quality Scores}
WQS correlates highly with raw reference counts across all six cohorts (0.74–0.80), indicating that reference quantity is a primary driver of WQS. Our Reference dimension correlates with WQS more weakly (0.45–0.63 in five cohorts; 0.79 in Brazilian oil paintings). This variance demonstrates the dimension's added value: it captures policy-compliance (by filtering out cross-wiki imports) and diversity, which are not covered by raw counts. While our metric cannot be completely decoupled from raw volume (showing high correlation in cohorts like Nigerian writers) the distinctly lower correlation found for well-documented cohorts like UK writers shows that WQS rewards mass reference count without distinguishing it from the validity, diversity, and density nuances captured by Reference.

\subsection{Specificity vs Claim Counts}
The Specificity dimension correlates highly with total claim counts in cohorts dominated by templated representations (Brazilian oil paintings: 0.91) and weakly in cohorts where editors curate distinguishing values (UK female writers: 0.19; UK male writers: 0.10). Intermediate cohorts (French oil paintings; Nigerian writers) fall in between (0.53–0.66). This shows that this dimension detects what it is designed to. In well-curated cohorts, Specificity captures editorial attention beyond raw volume, identifying when editors use rare, low-probability values rather than common template fillers.
The same pattern holds for WQS, which correlates with total claim counts more strongly than with Specificity (writers: WQS 0.66–0.74 vs Specificity 0.07–0.44; paintings: WQS 0.38–0.54 vs Specificity 0.26–0.34), with the correlation again negligible for UK writers. WQS does not adjust for templated, highly predictable claims but Specificity does.

Similarly, although Recoin is designed around property presence, it correlates most strongly with total claim counts in all contexts (0.57–0.64), except for Brazilian oil paintings where it drops to 0.20. Recoin generally fails to capture the nuances of our framework, with correlations to our six dimensions remaining low. It only reaches above 0.50 in a few domain-specific cases: Specificity in female Nigerian writers (0.53), and the External and Linguistic dimensions in UK writers (Female: 0.51 and 0.55; Male: 0.52 and 0.52).

\subsection{Connectivity and Reach: External and Linguistic Dimensions}
External and Linguistic dimensions correlate strongly with sitelinks for UK writers, where international recognition aligns with broad Wikipedia coverage, but decouple in other cohorts (External in particular for Nigerian writers, and both External and Linguistic for Brazilian paintings). This decoupling reflects an ontological distinction which the metric captures: locally recognised entities can have substantial external integration and linguistic accessibility without correspondingly broad cross-wiki coverage. Sitelinks alone treat such entities as poorly connected, while our framework does not.

\subsection{Qualifier Patterns}
The Qualifier dimension correlates near-perfectly with raw qualifier counts in all four writer cohorts (0.91–1.00), indicating that qualifier-permitting properties are uniformly used in those domains. For French oil paintings, the correlation drops to 0.52. This divergence highlights the context-awareness of the dimension. While cohorts with highly templated representations, such as Brazilian paintings, show a high correlation with raw counts, the metric captures nuances that simple volume misses in more complex cohorts. Qualifier use in the art domain diverges from the properties that structurally permit qualifiers. This decoupling can indicate either inconsistent qualifier use across the cohort or a graph structure that does not match domain editorial practice, both findings actionable for the relevant WikiProjects. Ultimately, these results confirm that while richness is related to completeness, it acts as a distinct quality dimension that extends beyond basic property presence.

\subsection{Linguistic Collinearity}
The Linguistic dimension shows near-perfect collinearity with raw language counts across all cohorts (0.91–0.99). In our case studies, languages with labels tend also to have descriptions and aliases, so the dimension's components scale together and the score reduces approximately to the label count. Despite this high correlation in these two specific cases, the metric is designed to handle edge cases in the wider Wikidata graph that raw counts fail on. The dimension is designed to capture variation in description quality and alias coverage that does not vary meaningfully in our cohorts. We expect the dimension to diverge from raw counts in contexts with more variable description quality across languages, but this remains to be empirically verified.

\section{Conclusion}

In this paper, we introduced SQuIREL, an entity richness framework to evaluate representational disparity in Wikidata. Moving beyond schema-level richness and standard completeness metrics, it deconstructs entity richness into six context-relative dimensions: Semantic Specificity, Qualifier Density, Reference Provenance, Internal Connectivity, External Network Connectivity, and Linguistic Accessibility, adapting Information Content theory from concepts to property-value pairs within a cohort. By applying this methodology to two case studies, oil paintings and writers, we demonstrated that representational disparity persists after context-relative normalisation and is not merely an artefact of cohort size. Our findings align with historical geographical skews favouring Global North entities (UK, France) over those in the Global South (Nigeria, Brazil). We also uncover intersectional patterns aggregate metrics obscure: in the UK, male writers are advantaged across most dimensions, while in Nigeria, gender gaps run in different directions, with female writers slightly outscoring male writers, possibly reflecting targeted editing initiatives aimed at improving representation of women from under-represented regions. Validation against the Weighted Quality Score, Recoin, sitelinks, and raw counts showed that the framework captures nuances these indicators miss. By keeping the six dimensions disaggregated, the framework provides the Wikidata community a complementary tool that identifies which dimension a cohort lacks, allowing editors to target interventions at specific gaps. Future work will focus on a public, browser-based tool for non-expert users and experiments on additional domains. We show in the appendix a structured methodology and worked examples on how the framework's outputs translate into actionable editorial recommendations.

\textbf{Limitations} 
Our framework surfaces representational disparity but cannot distinguish between editor inattention, source-material scarcity, and domain-specific modelling conventions as the underlying cause which requires complementary methods. Future mixed-methods research could pair this richness framework with external globally diverse archival sources. While our case studies were purposefully selected to test the metric across both human entities and physical artifacts of varying baseline volumes, we acknowledge that relying on only two case studies is a limitation. As it stands, the current framework serves as a proof of concept, and testing it on a wider range of domains (such as events, organisations, and geographic entities) will be an important focus of future work to strengthen our generalisability claim. The Linguistic dimension's collinearity with raw language counts in our cohorts (6.5) is a known limitation of the validation. The current weighting acts as a structural baseline for this proof-of-concept, but testing different calibrated weights based on per-language requirements are an avenue for future extension. 

\section*{Acknowledgments}
This work was supported by the UK Research and Innovation Centre for Doctoral Training in Safe and Trusted Artificial Intelligence\footnote{\url{https://www.safeandtrustedai.org}} [grant number EP/S023356/1].

\bibliographystyle{splncs04}
\bibliography{lib}

\newpage

\appendix

\section{SPARQL Queries}

\begin{figure}[!htbp]
\centering
\begin{tcolorbox}[
    colback=gray!10, 
    colframe=gray!40, 
    arc=3mm, 
    boxrule=1pt,
    title={Dimension 1: Semantic Specificity Queries},
    coltitle=black,  
    fonttitle=\bfseries     
]

\textbf{Query 1: Extract External ID property types}
\begin{verbatim}
    SELECT ?property WHERE {{
        VALUES ?item {{{entity_ids}}} .
        ?item ?p [] .
        ?property wikibase:claim ?p .
        ?property wikibase:propertyType wikibase:ExternalId . 
    }}
\end{verbatim}

\tcblower 

\textbf{Query 2: Extract values for specific properties}
\begin{verbatim}
    SELECT ?item ?property ?value WHERE {{
        VALUES ?property {{{property_ids}}}
        VALUES ?item {{{entity_ids}}}
        ?item ?property ?value .
    }}
\end{verbatim}
\end{tcolorbox}
\caption{SPARQL queries for filtering item statements by wikibase:ExternalId (top) and extracting values for a specific set of properties (bottom).}
\label{fig:combined_sparql_queries}
\end{figure}

\begin{figure}[!htbp]
\centering
\begin{tcolorbox}[
    colback=gray!10, 
    colframe=gray!40, 
    arc=3mm, 
    boxrule=1pt,
    title={Dimension 2: Qualifier Density Queries}, 
    coltitle=black,                  
    fonttitle=\bfseries              
]

\textbf{Query 1: Count allowed qualifiers in constraint rules}
\begin{verbatim}
    SELECT ?property (COUNT(?allowedQualifier) AS 
    ?allowedQualifierCount) 
    WHERE {{
        VALUES ?property {{{entity_ids}}}
        ?property p:P2302 ?constraintStatement .
        ?constraintStatement ps:P2302 wd:Q21510851 .  
        ?constraintStatement pq:P2306 ?allowedQualifier .  
    }}
    GROUP BY ?property
\end{verbatim}

\tcblower

\textbf{Query 2: Calculate total and distinct qualifier properties}
\begin{verbatim}
    SELECT ?item ?property ?statement (COUNT(?qualifierProperty) 
    AS ?qualifierCount) (COUNT(DISTINCT ?qualifierProperty) AS 
    ?qualifierdistinctCount)
    WHERE {{
        VALUES ?item {{{entity_ids}}} .
        ?item ?property ?statement .
                    
        FILTER (?property IN ({property_ids}))
        ?statement ?qualifierProperty ?qualifierValue .
        [] wikibase:qualifier ?qualifierProperty .
    }}
    GROUP BY ?item ?property ?statement
    ORDER BY DESC(?qualifierCount)
\end{verbatim}
\end{tcolorbox}
\caption{SPARQL queries for counting the number of allowed qualifiers defined in constraint rules (top) and calculating total and distinct qualifier properties attached to specific statement nodes (bottom). \textit{(P2302 - property constraint, P2306 - property qualifier, Q21510851 - allowed qualifiers constraint)}}
\label{fig:combined_qualifier_queries}
\end{figure}

\begin{figure}[!htbp]
\centering
\begin{tcolorbox}[
    colback=gray!10, 
    colframe=gray!40, 
    arc=3mm, 
    boxrule=1pt,
    title={Dimension 3: Reference Provenance Query},
    coltitle=black,                  
    fonttitle=\bfseries              
]

\textbf{Count statements and references excluding Wikimedia imports}
\begin{verbatim}
    SELECT ?item ?statementCount 
        (COALESCE(?refCount,0) AS ?referenceCount) 
        (COALESCE(?distinctRefCount,0) AS ?distinctreferenceCount)
        WHERE {{
            VALUES ?item {{{entity_ids}}} .
            {{
            SELECT ?item (COUNT(DISTINCT ?statement) AS 
            ?statementCount)
            WHERE {{
                VALUES ?item {{{entity_ids}}} .
                ?item ?p ?statement .
                ?statement wikibase:rank [] .
            }}
            GROUP BY ?item
        }}
          
        OPTIONAL {{
            SELECT ?item (COUNT(?reference) AS ?refCount) 
            (COUNT(DISTINCT ?reference) AS ?distinctRefCount)
            WHERE {{
                VALUES ?item {{{entity_ids}}} .
                ?item ?p ?statement .
                ?statement wikibase:rank [] .
                ?statement prov:wasDerivedFrom ?reference .
              
                MINUS {{ ?reference pr:P143 [] . }}
            }}
            GROUP BY ?item
        }}
    }}
\end{verbatim}
\end{tcolorbox}
\caption{SPARQL query to count the total number of statements and references for specific Wikidata entities, excluding references that were copied from other Wikimedia projects.}
\label{fig:query_related_properties}
\end{figure}

\begin{figure}[!htbp]
\centering
\begin{tcolorbox}[
    colback=gray!10, 
    colframe=gray!40, 
    arc=3mm, 
    boxrule=1pt,
    title={Dimension 4: Internal Connectivity Query},
    coltitle=black,                  
    fonttitle=\bfseries              
]

\textbf{Count incoming links to target entities}
\begin{verbatim}
    SELECT ?value (COUNT(*) AS ?count)
    WHERE {{
        VALUES ?value {{{entity_ids}}} .
            ?subject ?p ?value .
            FILTER(STRSTARTS(STR(?p), STR(wdt:)))
            FILTER(STRSTARTS(STR(?subject), 
            "http://www.wikidata.org/entity/Q"))
    }}
    GROUP BY ?value 
\end{verbatim}
\end{tcolorbox}
\caption{SPARQL query to count the total number of incoming links from other Wikidata items to a specific set of target entities.}
\label{fig:incoming_links}
\end{figure}

\begin{figure}[!htbp]
\centering
\begin{tcolorbox}[
    colback=gray!10, 
    colframe=gray!40, 
    arc=3mm, 
    boxrule=1pt,
    title={Dimension 5: External Connectivity Queries},
    coltitle=black,                  
    fonttitle=\bfseries              
]

\textbf{Query 1: Count and rank external identifier properties}
\begin{verbatim}
    SELECT ?property (COUNT(?property) AS ?propertyCount) 
    WHERE {{
        VALUES ?item {{{entity_ids}}} .
        ?item ?p [] .
        ?property wikibase:claim ?p .
        ?property wikibase:propertyType wikibase:ExternalId . 
    }}
    GROUP BY ?property
    ORDER BY DESC(?propertyCount)
\end{verbatim}

\tcblower 

\textbf{Query 2: Count usage of specific properties on entities}
\begin{verbatim}
    SELECT ?item (COUNT(?property) AS ?propertyCount) 
    WHERE {{
        VALUES ?item {{{entity_ids}}} .
        VALUES ?property {{{property_ids}}} .
        ?item ?property [] .
    }}
    GROUP BY ?item
\end{verbatim}
\end{tcolorbox}
\caption{SPARQL queries to count and rank external identifier properties (top) and count the total usage of a specific set of properties on provided entities (bottom).}
\label{fig:combined_property_queries}
\end{figure}

\begin{figure}[!htbp]
\centering
\begin{tcolorbox}[
    colback=gray!10, 
    colframe=gray!40, 
    arc=3mm, 
    boxrule=1pt,
    title={Dimension 6: Linguistic Accessibility Queries},
    coltitle=black,                  
    fonttitle=\bfseries              
]

\textbf{Query 1: Extract languages of translated labels}
\begin{verbatim}
    SELECT DISTINCT ?item ?lang
    WHERE {{
        VALUES ?item {{{entity_ids}}} 
        ?item rdfs:label ?label .
        BIND(LANG(?label) AS ?lang)
    }}
    ORDER BY ?item ?lang
\end{verbatim}

\vspace{0mm}
\noindent\tikz\draw[gray!40, dashed, line width=1pt] (0,0) -- (\linewidth,0);
\vspace{0mm}

\textbf{Query 2: Count aliases by language}
\begin{verbatim}
    SELECT ?item ?lang (COUNT(?alias) AS ?aliasCount)
    WHERE {{
        VALUES ?item {{{entity_ids}}}
        ?item skos:altLabel ?alias .
    }}
    GROUP BY ?item (LANG(?alias) AS ?lang)
    ORDER BY DESC(?aliasCount) 
\end{verbatim}

\vspace{0mm}
\noindent\tikz\draw[gray!40, dashed, line width=1pt] (0,0) -- (\linewidth,0);
\vspace{0mm}

\textbf{Query 3: Extract descriptions and calculate lengths}
\begin{verbatim}
    SELECT ?item ?lang ?description (STRLEN(?description) AS 
    ?char_count)
    WHERE {{
        VALUES ?item {{{entity_ids}}}
        ?item schema:description ?description .
        BIND(LANG(?description) AS ?lang)
    }}
\end{verbatim}
\end{tcolorbox}
\caption{SPARQL queries to acquire the languages of translated labels (top), count the number of aliases by language (middle), and extract descriptions alongside their character lengths across available languages (bottom).}
\label{fig:multilingual_data_queries}
\end{figure}

\clearpage

\section{Case Study 1: Grouped}

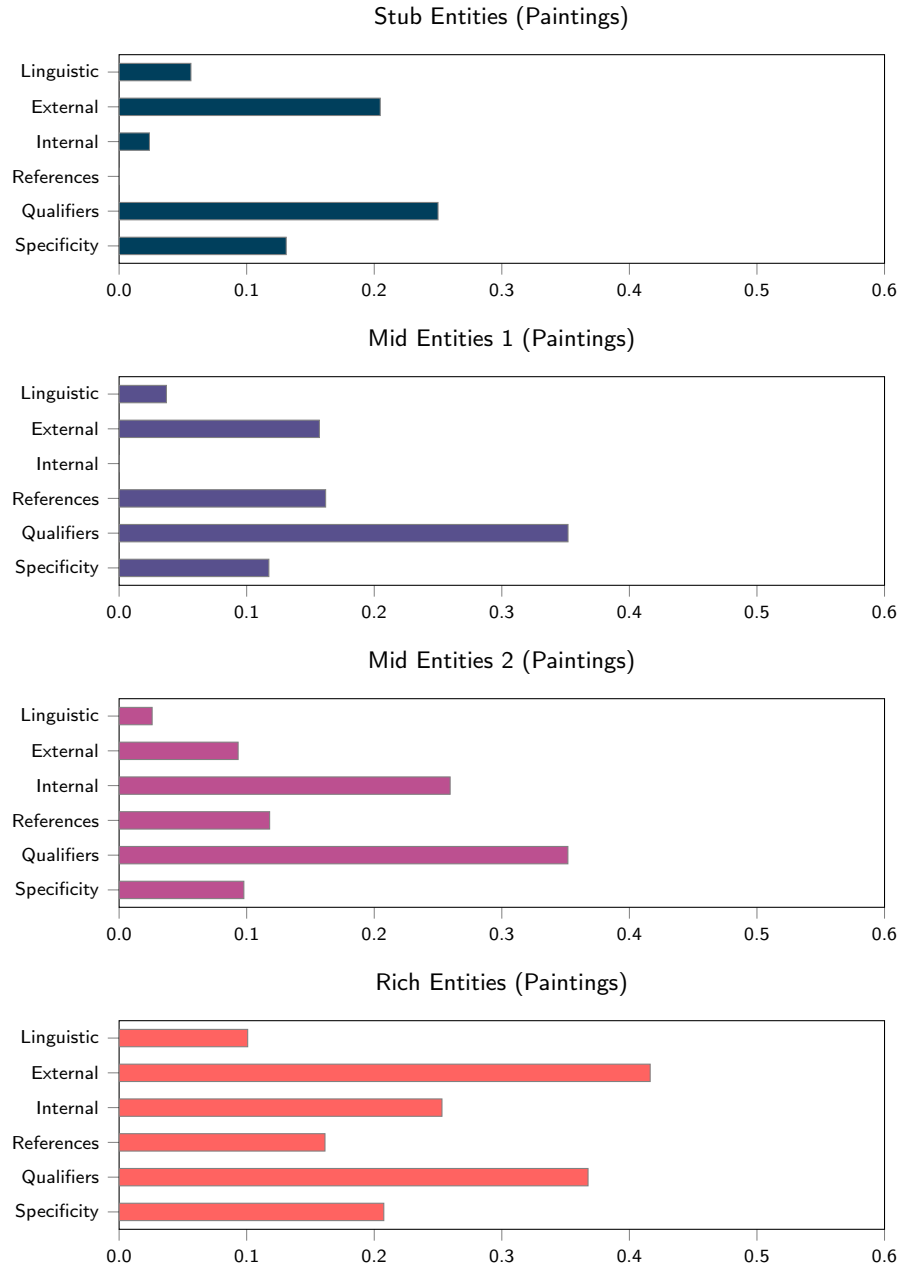
\begin{figure}[!htbp]
\centering
{\sffamily\sansmath
\begin{tikzpicture}
\begin{groupplot}[
    group style={group size=1 by 4, vertical sep=1.5cm},
    scale only axis,
    width=0.83\linewidth,
    height=2.76cm,
    every axis plot/.append style={xbar, bar width=0.5},
    xmin=0, xmax=0.6,
    ymin=-0.5, ymax=5.5,
    y dir=reverse,
    ytick={0,1,2,3,4,5},
    yticklabels={Linguistic, External, Internal, References, Qualifiers, Specificity},
    xtick={0,0.1,0.2,0.3,0.4,0.5,0.6},
    xticklabels={0.0,0.1,0.2,0.3,0.4,0.5,0.6},
    tick label style={font=\scriptsize},
    title style={font=\small},
    tick align=outside,
    xtick pos=bottom, ytick pos=left,
]
\nextgroupplot[title={Stub Entities (Paintings)}]
\addplot[fill=stub, draw=gray] coordinates
    {(0.0563,0) (0.2047,1) (0.0238,2) (0,3) (0.2500,4) (0.1310,5)};
\nextgroupplot[title={Mid Entities 1 (Paintings)}]
\addplot[fill=mid1, draw=gray] coordinates
    {(0.0371,0) (0.1570,1) (0,2) (0.1619,3) (0.3519,4) (0.1174,5)};
\nextgroupplot[title={Mid Entities 2 (Paintings)}]
\addplot[fill=mid2, draw=gray] coordinates
    {(0.0259,0) (0.0933,1) (0.2595,2) (0.1180,3) (0.3518,4) (0.0978,5)};
\nextgroupplot[title={Rich Entities (Paintings)}]
\addplot[fill=rich, draw=gray] coordinates
    {(0.1008,0) (0.4163,1) (0.2531,2) (0.1614,3) (0.3676,4) (0.2075,5)};
\end{groupplot}
\end{tikzpicture}}
\caption{Metric breakdowns by entity richness level (Stub, Mid, Rich) for paintings.}
\label{fig:painting_grouped}
\end{figure}

\newpage

\section{Case Study 1: Gini Scores}

\begin{figure*}[!htbp]
\centering
\includegraphics[width=1\textwidth]{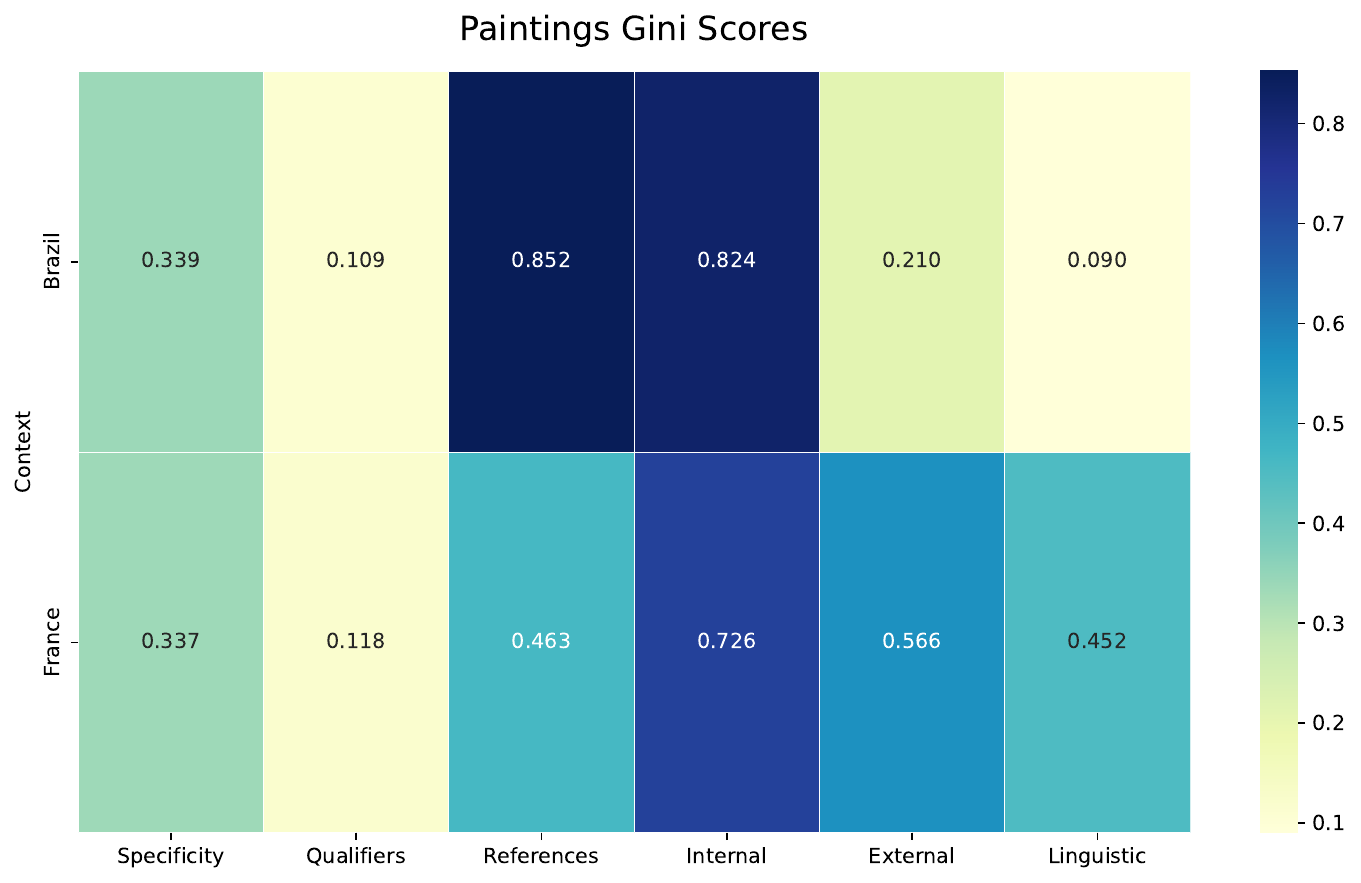} 
\caption{Gini scores for painting entities across the six dimensions.}
\label{fig:painting_gini}
\end{figure*}

\newpage

\section{Case Study 2: Grouped}

\begin{figure}[!htbp]
\centering
{\sffamily\sansmath
\begin{tikzpicture}
\begin{groupplot}[
    group style={group size=1 by 3, vertical sep=1.5cm},
    scale only axis,
    width=0.83\linewidth,
    height=3.6cm,
    every axis plot/.append style={xbar, bar width=0.5},
    xmin=0, xmax=1,
    ymin=-0.5, ymax=5.5,
    y dir=reverse,
    ytick={0,1,2,3,4,5},
    yticklabels={Linguistic, External, Internal, References, Qualifiers, Specificity},
    xtick={0,0.2,0.4,0.6,0.8,1},
    xticklabels={0.0,0.2,0.4,0.6,0.8,1.0},
    tick label style={font=\scriptsize},
    title style={font=\small},
    tick align=outside,
    xtick pos=bottom, ytick pos=left,
]
\nextgroupplot[title={Stub Entities (Writers)}]
\addplot[fill=stub, draw=gray] coordinates
    {(0.0386,0) (0.2469,1) (0.0600,2) (0.2995,3) (0,4) (0.1233,5)};
\nextgroupplot[title={Mid Entities (Writers)}]
\addplot[fill=mid1, draw=gray] coordinates
    {(0.0630,0) (0.4523,1) (0.1298,2) (0.3891,3) (0.2209,4) (0.1353,5)};
\nextgroupplot[title={Rich Entities (Writers)}]
\addplot[fill=mid2, draw=gray] coordinates
    {(0.1367,0) (1.0000,1) (0.2998,2) (0.4294,3) (0.2174,4) (0.1373,5)};
\end{groupplot}
\end{tikzpicture}}
\caption{Metric breakdowns by entity richness level (Stub, Mid, Rich) for writers.}
\label{fig:writer_grouped}
\end{figure}
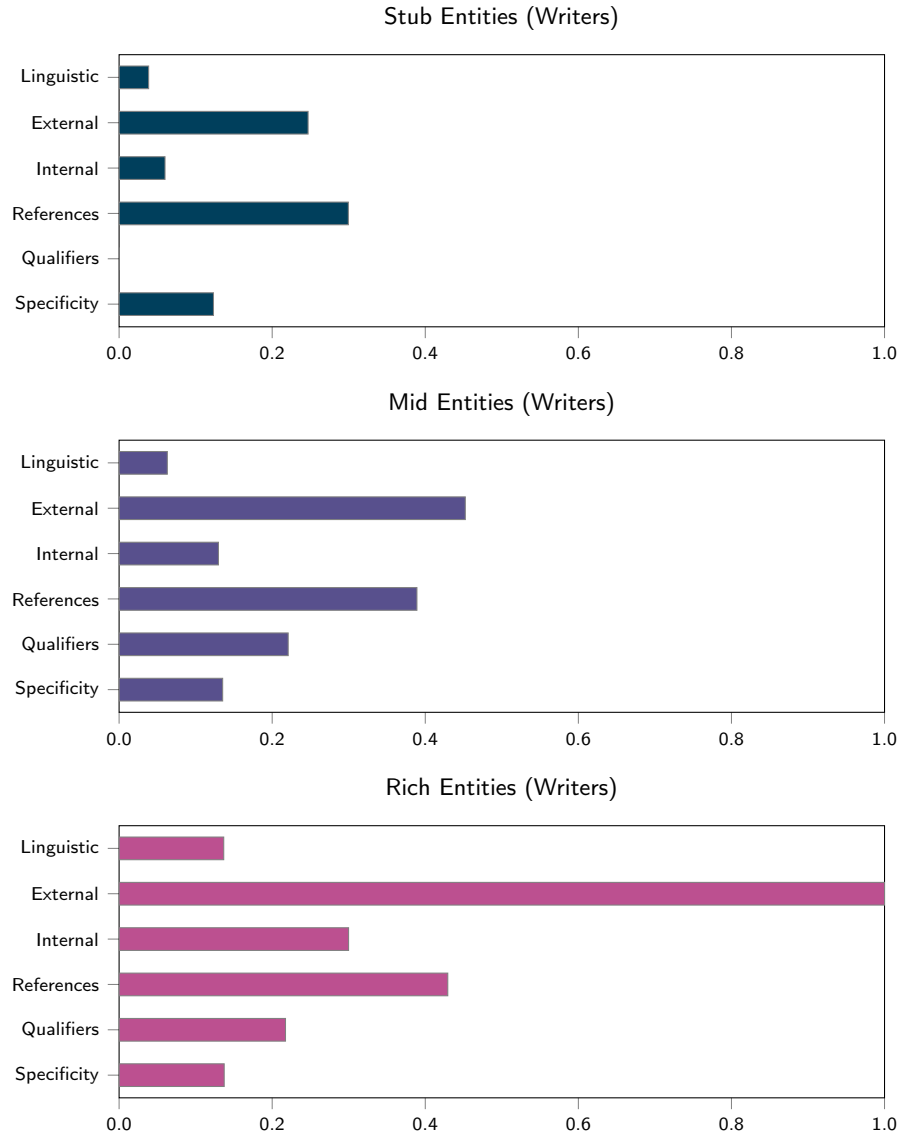

\newpage

\section{Case Study 2: Gini Scores}

\begin{figure*}[!htbp]
\centering
\includegraphics[width=1\textwidth]{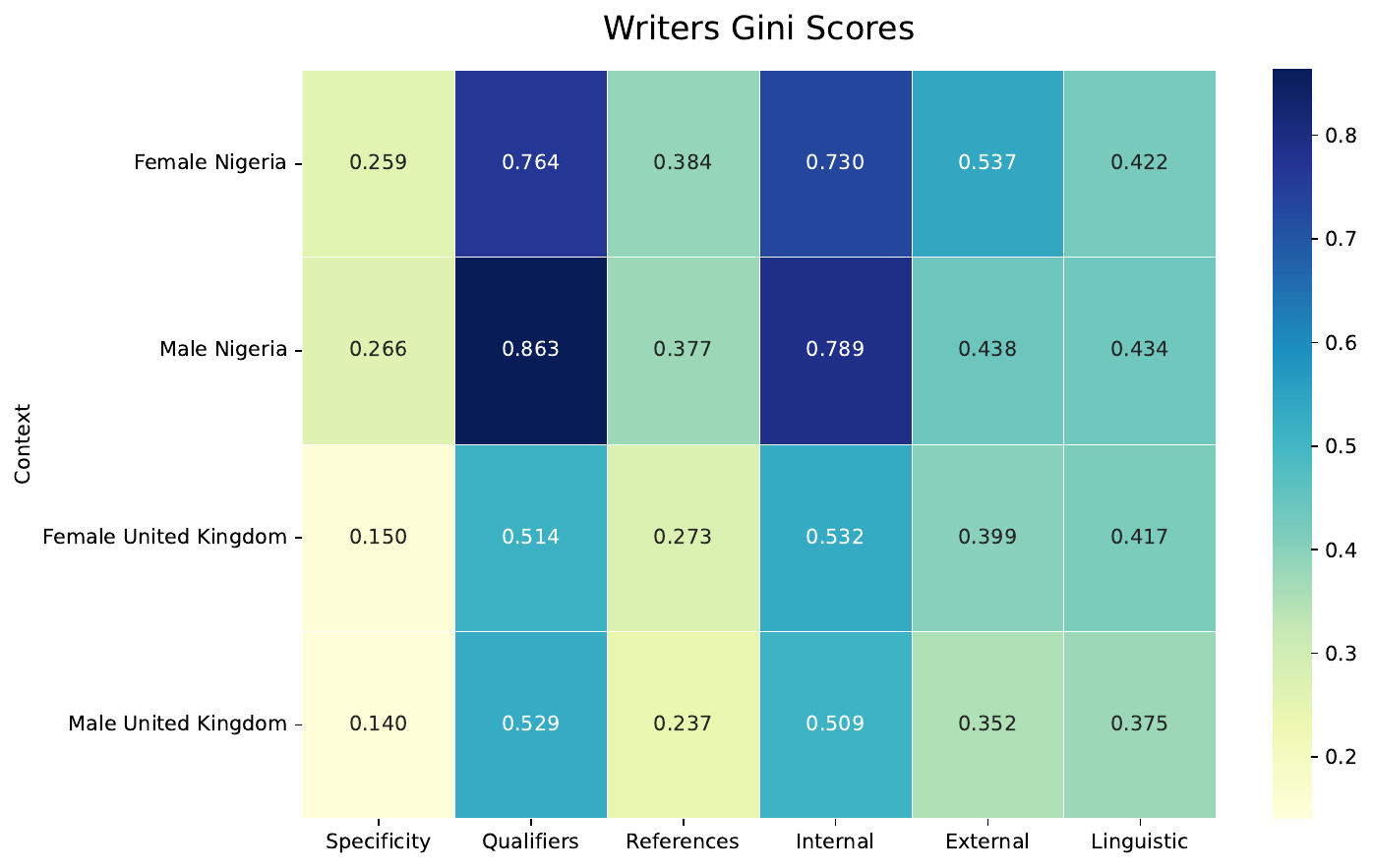} 
\caption{Gini scores for writer entities across the six dimensions.}
\label{fig:writer_gini}
\end{figure*}

\newpage

\section{Case Study 2: Pairwise Comparisons}

\begin{table}[!htbp]
    \centering
    \resizebox{0.9\textwidth}{!}{
    \begin{tabular}{l l r c l}
        \toprule
        \textbf{Dimension} & \textbf{Comparison} & \textbf{Adj. $p$-value} & \textbf{Effect Size ($r$)} & \textbf{Higher Median} \\
        \midrule
        \textbf{Specificity} & F-NG vs M-NG & .396 & .059 & - \\
         & F-NG vs F-UK & $< .001$ & .184 & F-UK \\
         & F-NG vs M-UK & $< .001$ & .157 & M-UK \\
         & M-NG vs F-UK & $< .001$ & .254 & F-UK \\
         & M-NG vs M-UK & $< .001$ & .228 & M-UK \\
         & F-UK vs M-UK & $< .001$ & .043 & F-UK \\
        \midrule
        \textbf{Qualifiers} & F-NG vs M-NG & $< .001$ & .133 & Tie \\
         & F-NG vs F-UK & $< .001$ & .298 & F-UK \\
         & F-NG vs M-UK & $< .001$ & .307 & M-UK \\
         & M-NG vs F-UK & $< .001$ & .437 & F-UK \\
         & M-NG vs M-UK & $< .001$ & .439 & M-UK \\
         & F-UK vs M-UK & .129 & .024 & - \\
        \midrule
        \textbf{References} & F-NG vs M-NG & .039 & .087 & F-NG \\
         & F-NG vs F-UK & $< .001$ & .202 & F-UK \\
         & F-NG vs M-UK & $< .001$ & .271 & M-UK \\
         & M-NG vs F-UK & $< .001$ & .318 & F-UK \\
         & M-NG vs M-UK & $< .001$ & .397 & M-UK \\
         & F-UK vs M-UK & $< .001$ & .065 & M-UK \\
        \midrule
        \textbf{Internal} & F-NG vs M-NG & $< .001$ & .102 & Tie \\
         & F-NG vs F-UK & $< .001$ & .324 & F-UK \\
         & F-NG vs M-UK & $< .001$ & .405 & M-UK \\
         & M-NG vs F-UK & $< .001$ & .429 & F-UK \\
         & M-NG vs M-UK & $< .001$ & .504 & M-UK \\
         & F-UK vs M-UK & $< .001$ & .098 & Tie \\
        \midrule
        \textbf{External} & F-NG vs M-NG & $< .001$ & .161 & M-NG \\
         & F-NG vs F-UK & $< .001$ & .676 & F-UK \\
         & F-NG vs M-UK & $< .001$ & .745 & M-UK \\
         & M-NG vs F-UK & $< .001$ & .632 & F-UK \\
         & M-NG vs M-UK & $< .001$ & .710 & M-UK \\
         & F-UK vs M-UK & $< .001$ & .155 & M-UK \\
        \midrule
        \textbf{Linguistic} & F-NG vs M-NG & $< .001$ & .324 & F-NG \\
         & F-NG vs F-UK & $< .001$ & .477 & F-UK \\
         & F-NG vs M-UK & $< .001$ & .641 & M-UK \\
         & M-NG vs F-UK & $< .001$ & .640 & F-UK \\
         & M-NG vs M-UK & $< .001$ & .794 & M-UK \\
         & F-UK vs M-UK & $< .001$ & .136 & M-UK \\
        \bottomrule
    \end{tabular}
    }
    \caption{Pairwise Comparisons (Mann-Whitney U) for Writer Scores. \textit{Note: F-NG = Female Nigeria, M-NG = Male Nigeria, F-UK = Female United Kingdom, M-UK = Male United Kingdom. P-values are Bonferroni-adjusted.}}
    \label{tab:posthoc_stats}
\end{table}

\newpage

\section{Correlation Matrix}

\begin{figure*}[!htbp]
\centering
\includegraphics[width=0.9\textwidth]{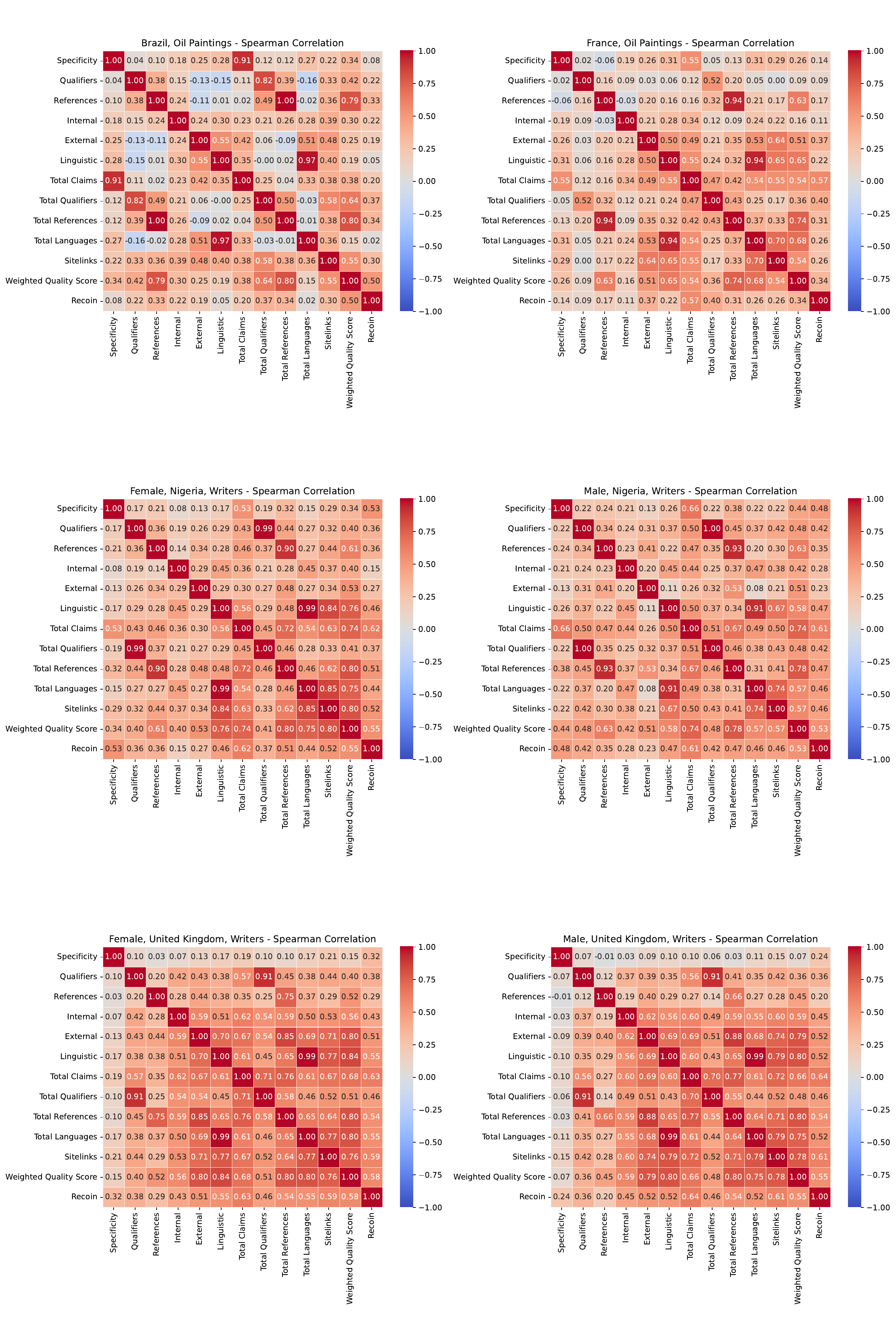} 
\caption{Spearman correlation matrices of richness dimensions and quality scores across all writer (by gender and nationality) and painting (by country) sub-groups.}
\label{fig:corr_matrix}
\end{figure*}

\newpage

\section{Documentation Framework}

The following is an outline and explanation of the documentation framework that can be used to document the results of the six dimensions of the proposed framework when used. \newline

\noindent \textbf{Define Context Constants:}
State the constant property-value pairs that were used to gather items to populate the context. \newline

\noindent \textbf{Comparison Cohorts:}
State the varied property-value pairs that define the focal and reference cohorts.\newline

\noindent \textbf{Main Bias Indicator(s):}
From the results of the six richness dimensions, what are the main indicators of bias when comparing the focal cohort to the reference cohort?\newline

\noindent \textbf{Editorial Gap:}
Is the bias caused by a skew in editor attention or behaviour? Does the focal cohort have a significant deficit in enriched items? Does the focal cohort have higher Gini scores due to a lack of editor attention, community focus, or organised WikiProjects compared to the reference cohort?\newline

\noindent \textbf{Source Material Gap:}
Is the bias caused by a lack of verifiable references and source material? Does the focal cohort have lower Reference scores and more stub items? Is the disparity a reflection of external reality? Do the reference cohort's items simply have more published, verifiable literature globally available than the focal cohort's items?\newline

\noindent \textbf{Data Model Gap:}
Is the bias caused by the mismatch between the data model and the epistemic reality of the focal cohort? Does the high Qualifier or Internal Link Gini suggest that Wikidata's established schema is poorly suited for the focal cohort's context? Are editors forced to leave items bare because the culturally or contextually correct properties do not exist? \newline

\noindent \textbf{Bias Check:}
Does the high inequality indicate actual bias, or a natural structural hierarchy within this specific domain (e.g., a few ``hub'' items naturally hoarding links while other valid items have few)? \newline

\noindent \textbf{Targeted Recommendation(s):}
We split the target recommendations into four categories that can be utilised when applicable.
\begin{itemize}
    \item \textbf{Parity:} Narrow the gap in profile distribution between the focal and reference cohorts
    \item \textbf{Floor Raising:} Reduce the stub entities
    \item \textbf{Investigation:} The cause of bias is unknown and should be investigated
    \item \textbf{No internal action:} Bias is external/epistemological; Wikidata intervention will not solve root cause
\end{itemize}

\noindent \textbf{Actionable Goal(s):}
Here actionable goals can be listed to create an itinerary of interventions for the Wikidata community to potentially tackle.\newline

\section{Case Study 1: Oil Paintings (Brazil vs France)}

\noindent \textbf{Context constants:} P31=Q3305213 (painting), P186=Q296955 (oil paint) \newline

\noindent \textbf{Reference Cohort:} France (Q142), n=901
\noindent \textbf{Focal Cohort:} Brazil (Q155), n=852 \newline

\noindent \textbf{Main bias indicators:}
\begin{itemize}
    \item Stub cluster: 48\% Brazil vs 0\% France
    \item Rich cluster: 13\% Brazil vs 20\% France
    \item References: largest distributional gap, Brazil clusters near zero
    \item Internal Connectivity: Brazil clusters near zero, France broadly distributed
\end{itemize}

\noindent \textbf{Editorial gap:} Strong. Reference Gini 0.85 (Brazil) vs 0.46 (France). For Brazil this could indicate inconsistent attention and not systematic curation. \newline

\noindent \textbf{Source material gap:} Contributing, not sufficient. External scores show institutional records exist. Source material available but not being surfaced. \newline

\noindent \textbf{Data model gap:} None identified. Qualifier distributions similar in shape, Brazil Specificity median slightly higher than France. Schema accommodates Brazil works. \newline

\noindent \textbf{Bias check:}

\begin{itemize}
    \item Internal Gini high in both (0.82 BR / 0.73 FR), partly natural scale-free structure
    \item Reference Gini divergence (0.85 vs 0.46), not natural, no domain reason
    \item 48\% vs 0\% stub asymmetry, too extreme for chance
    \item Conclusion: real disparity, not domain artefact
\end{itemize}

\noindent \textbf{Recommendations:}
\begin{itemize}
    \item Floor Raising (primary): target 48\% stub cluster
    \item Parity (secondary): References specifically; Specificity/External/Linguistic already at parity
    \item Investigation: External-vs-Internal disconnect (institutional recognition without graph integration)
    \item No internal action: not applicable
\end{itemize}

\noindent \textbf{Goals:}

\begin{itemize}
    \item Reduce stub entities for Brazilian paintings
    \item Engage with specific communities
    \item Audit External-high / Internal-low Brazilian paintings, add incoming links
\end{itemize}

\section{Case Study 2: Writers (UK vs Nigeria, female vs male)}

\noindent \textbf{Context constants:} P31=Q5 (human), P106=Q36180 (writer) \newline

\noindent \textbf{Reference Cohort 1:} UK female (Q145, Q6581072), n=4,649
\noindent \textbf{Reference Cohort 2:} UK male (Q145, Q6581097), n=7,996
\noindent \textbf{Focal Cohort 1:} Nigeria female (Q1033, Q6581072), n=426
\noindent \textbf{Focal Cohort 2:} Nigeria male (Q1033, Q6581097), n=1,396
\newline

\noindent \textbf{Main bias indicators:}
\begin{itemize}
    \item Citizenship dominates variance
    \item Largest gap: External Connectivity
    \item Rich cluster: 1\% Nigeria (both genders) vs 15\% UK female / 24\% UK male
    \item Stub cluster: 66\% Nigeria female / 80\% NG male vs 33\% UK female / 32\% UK male
    \item Gender direction reverses across citizenship: UK male > UK female on References/External/Linguistic; Nigeria female > Nigeria male on References
\end{itemize}

\noindent \textbf{Editorial gap:}
Citizenship axis: Strong. Nigeria Gini larger than UK Gini across Specificity, Qualifiers, References, Internal. Nigeria male extreme imbalance (Qualifier Gini 0.86, Internal 0.79).
Gender axis (Nigeria): less percentage of stub entities for female. Possible explanation could be targeted debiasing initiatives (Art+Feminism, AfroCROWD, Whose Knowledge?) producing curated entries for small Nigeria female cohort. Hypothesis, not confirmed. \newline

\noindent \textbf{Source material gap:} 
Possible substantial contributor to External gap, as identifier systems typically carry geographic bias and under-index African literary production. \newline

\noindent \textbf{Data model gap: }
Possible, secondary. Writer schema not obviously mismatched but could be contributor due to less coverage of non-Western literary genres.\newline

\noindent \textbf{Bias check:}

\begin{itemize}
    \item Citizenship bias is very likely. Consistent across dimensions, cluster composition, Gini scores
    \item Gender within Nigeria: small cohort (n=426) introduces variability.
    \item Gender within UK: consistent with prior gender-bias literature.
\end{itemize}

\noindent \textbf{Recommendations:}

\begin{itemize}
    \item Floor Raising (primary): 66-80\% Nigeria stub
    \item Parity (secondary): Specificity, Qualifier dimensions. Not External as it is due to upstream bias.
    \item Investigation: Nigeria female-over-male References finding. Cross-reference edit history with initiative timelines.
    \item No internal action: External Connectivity gap rooted in upstream identifier bias. Should be documented.
\end{itemize}

\noindent \textbf{Goals:}

\begin{itemize}
    \item Reduce stub entities for Nigerian writers
    \item Property coverage audit and propose new properties where warranted
    \item Replicate gender direction finding, on other writers cohorts
    \item Document upstream identifier bias as separate report; not a Wikidata-internal goal
\end{itemize}

\end{document}